\pdfoutput=1

\documentclass[11pt]{article}

\usepackage[]{EMNLP2022}

\usepackage{times}
\usepackage{latexsym}

\usepackage[T1]{fontenc}

\usepackage[utf8]{inputenc}

\usepackage{microtype}

\usepackage{inconsolata}
\usepackage{subcaption}
\usepackage{graphicx}
\usepackage{placeins}

\title{Individuation in Neural Models with and without Visual Grounding}

\author{
Alexey Tikhonov\\
Inworld.AI, \\
Berlin, Germany\\
\texttt{altsoph@gmail.com}\And
Lisa Bylinina\\
Utrecht University\\
Utrecht, the Netherlands\\
\texttt{e.g.bylinina@uu.nl}\AND
Ivan P. Yamshchikov\\
CAIRO,
Technical University of Applied Sciences W{\"u}rzburg-Schweinfurt\\
W{\"u}rzburg, Germany\\
\texttt{ivan.yamshchikov@thws.de}
}

\begin{document}
\maketitle
\begin{abstract}
We show differences between a language-and-vision model CLIP and two text-only models — FastText  and SBERT — when it comes to the encoding of individuation information. We study latent representations that CLIP provides for substrates, granular aggregates, and various numbers of objects. We demonstrate that CLIP embeddings capture quantitative differences in individuation better than models trained on text-only data. Moreover, the individuation hierarchy we deduce from the CLIP embeddings agrees with the hierarchies proposed in linguistics and cognitive science.
\end{abstract}

\section{Introduction}

Recent results in multimodal\footnote{For a detailed review of various aspects of multimodal machine learning, we address the reader to \cite{zhang2020multimodal}} vision and language (V\&L) models lead to intriguing research questions. For instance, one exciting research direction would be to search for the synergistic effects of multimodality. So far, to the best of our knowledge, no definitive finds were made on this front, despite the growing body of research on V\&L model evaluation. For example, \citet{parcalabescu2021valse} provide a benchmark to assess the visual grounding capabilities of V\&L models. The authors conclude that current models have difficulty addressing most phenomena that require models to ground linguistic information in the visual modality.  \citet{thrush2022winoground} present a benchmark for visio-linguistic compositional reasoning and also find that none of the modern V\&L models does much better than chance.

This paper presents the first potential case of such visio-linguistic synergies. Namely, it studies the phenomenon of {\em individuation} and how V\&L models represent objects -- that is, how they distinguish objects from substances and how they track objects and their quantity in sets that contain more than one object. We demonstrate that CLIP's \cite{radford2021learning} latent representations have properties that differ from those of the models that use only textual data. Moreover, this emergent property seems to agree with individuation scales proposed earlier by linguists and cognition researchers.

\section{Individuation}


Individuation is generally understood as basic principles that guide the distinction between objects and substances, as well as the distinction between a single object and multiple objects. Individuation is not limited to visual modality -- it applies cross-modally to stimuli of any kind. Here, we will only focus on visual individuation and its relation to the linguistic properties of corresponding words. 
This section summarizes the main relevant findings on individuation from cognitive science and linguistics.


Operationally, individuation can be probed along two axes: 1) the quantity axis; 2) the object axis. The former corresponds to distinguishing and tracking individual objects as their quantity increases. The latter is the dependency between the individual properties of an object and its permeability as an object rather than a substance. We will now overview these two aspects of individuation -- first, in cognition, then in language.

\subsection{Individuation and Cognition}

\noindent {\bf The quantity axis}. The human ability to perceive, identify, track, and count objects generally decreases as the number of objects in a scene increases \cite{feigenson2004core,dehaene2011number,10.3389/fnhum.2011.00150}. The cognitive basis of this observation is complex. In particular, two relevant cognitive systems have been identified: the object tracking system (OTS) and the approximate number system (ANS) \cite{carey1998knowledge,spaepen2011number, spelke2011natural}. OTS is active when the number of objects to track is low, typically under 3 or 4. OTS tracks each object individually and represents the exact quantity of objects in a scene. ANS, on the contrary, does not construct individual object representations and does not track the exact quantity of objects. In particular, in a cardinality comparison task in which two sets of arbitrary objects are given, the ability to tell which of the two sets has higher cardinality depends on the ratio between the sets' cardinalities. In pre-verbal infants, this ratio can be around 1:2, but it decreases somewhat with development \cite{HYDE2010647}.

Summing up, the human ability to represent objects and their quantity is not stable across quantities, with sharp contrast at the edge of OTS, and is ratio-sensitive in the ANS domain. For example, ten versus fifteen would be more distinguishable than twenty versus twenty-five even though the absolute difference between the cardinalities is the same; see \cite{starkey1980perception}. Both systems are non-linguistic since they are present in pre-verbal humans.

\vspace*{.5ex}
\noindent {\bf The object axis}. Humans organize their visual space into objects vs. substances very early in life, well in the pre-verbal stage of their development. \cite{spelke1990principles} identifies the basic principles of such an organization as Cohesion, Boundedness, Rigidity, and No Action at a Distance. Objects defined against these principles are called `Spelke objects.' Such objects tend to be connected, non-overlapping, with constant spacial characteristics when moving and only affecting each other when in contact. This is not exactly the same notion of an object as found in adults: for example, under these principles, a horseman riding a horse would be considered one object with the horse. 

Individuation principles develop and change during the lifetime, but the most drastic changes happen around the first year and coincide with language acquisition breakthroughs. Knowledge about linguistic labels for classes of objects has been argued to be used in individuation at this stage \cite{xu2007sortal}. Still, the causal relation between linguistic milestones and the changes in individuation strategies is under debate. For example, it is hard to disentangle linguistic factors from the rapid accumulation of world knowledge happening in the same period. For a deeper discussion of these factors, see \cite{gentner2001individuation}, who also suggests a cognitive hierarchy of individuation as a development of `Spelke object' principles:

\begin{center}
humans < animals < vehicles < small mobile objects < complex structurally cohesive objects < amorphous 
\end{center}

\subsection{Individuation and Language}

Natural language shows systematic distinctions with respect to both the quantity and the object axes. These distinctions can be linked to the organization of the corresponding cognitive systems. 

\noindent{\bf The quantity axis}. Distinctions in the representation of different quantities in language grammar manifest themselves mainly in two domains: 1) number morphology; 2) morphosyntax of constructions with numerals.

In languages like English, morphological number distinctions give rise to a split between one object ({\it book}) and a higher number of objects ({\it book-s}; however, plural nouns can refer to singular objects as well, see \citealt{spector2007aspects,zweig2009number}). Some languages also have a dual grammatical number as part of the nominal number inventory, making the 1 vs. 2 vs. >2 quantity distinction in the number domain (Slovenian, Arabic, etc.). Few languages also have the trial number form (e.g., some Austronesian languages and Austronesian-influenced creoles) and paucal number form referring to a 'small' number of objects (e.g., some Oceanic languages). It's debated whether there are languages with the quadral number form. Higher grammaticalized number distinctions don't exist in natural language -- for instance, there is no morphological affix as part of the grammatical number category that would mean 'exactly 7' or '15 or more'. Within the space of existing number distinctions, the higher the number line, the rarer the distinction. For example, the trial number form is quite rare typologically; dual number form is more frequent but rarer than a system with just the singular vs. plural distinction. A number hierarchy supports this observation: if a language has some number form, it also has all the number forms to the left of it \cite{croft1990typology,corbett2000number}: 

\begin{center}
    singular < plural < dual < paucal/trial
\end{center}

Thus, quantity distinctions built into language grammar through number marking show up exclusively on the lower side of the number line, roughly in the subitizing/OTS domain. Still, even within this domain, different quantities are not equally distinguished -- the lower, the more prominent.

Constructions with numerals ({\it five books} etc.) communicate precise quantities. The morphosyntax of such constructions varies somewhat depending on the quantity encoded by the numeral -- in English, for example, numeral {\it one} combines with singular nouns ({\it one book}), while higher numerals combine with plural ({\it seven books}). This is not universally true (e.g., in Turkic languages, all numerals combine with nouns in singular form), but more importantly, in more morphologically rich languages than English, a variety of grammatical distinctions is made between different quantities in this domain. For instance, in Russian, numeral {\it two} agrees with the noun in gender and case, while {\it three} agrees only in case (see \citealt{MPMD6Q_2019} for data on the grammatical typology of numerals). The generalization is, again, that, like with number marking, low quantities systematically receive special grammatical treatment in numeral constructions: very few systematic distinctions are made above 3-4, and even within this range -- the lower the quantity, the more distinct it is from other quantities, grammatically. 

\vspace*{.5ex}
\noindent{\bf The object axis}. Linguistic individuation is most often discussed in the context of the mass vs. count distinction in nouns. This distinction roughly separates entities that are construable as individuatable from those that are not and comes in a variety of specific linguistic behaviors, some of which we list below (see \citealt{mufwene1981non,wierzbicka1985oats} a.o.).

\begin{itemize}
    \item Pluralization: Count nouns allow for plural marking, mass nouns don't ({\it books} vs. *{\it rices});
    \item Numerals: Count nouns allow for numeral modification, mass nouns don't ({\it three books} vs. {\it three rices});
    \item Count quantifiers: Count nouns combine with quantifiers {\it many}/{\it several}, mass nouns don't ({\it several books} vs ??{\it several rices});
    \item Extent quantifiers: Count nouns don't combine with quantifiers {\it much} / {\it few}, mass nouns do (*{\it much book(s)} vs {\it much rice}).
\end{itemize}

\cite{grimm2012number} provides a much more extensive inventory of tests relevant to the same underlying distinction and extending beyond English. These tests suggest a coherent space of linguistic distinctions that gives rise to a very detailed hierarchy (for earlier versions of this hierarchy, see \cite{allan1980nouns, comrie1989language, croft1990typology}: 

\begin{center}
    liquids < foodstuffs < granular aggregate < vegetation/cereals/fruits $\leq$ insects < small animals < pair/grouped body parts $\leq$ middle-sized animals < types of people < individuals
\end{center}

\cite{grimm2012number} suggests that some of the details in the hierarchy above might be rooted in the specifics of the language sample used in his work. Thus, a simplification/generalization is proposed as follows:

\begin{center}
    liquids/substances < granular aggregates < collective aggregates < individuals
\end{center}

The above suggests that many clues in language alone can help deduce a hierarchy parallel to the non-linguistic cognitive hierarchy of individuation. But many such linguistic clues are language-specific and don't surface in, for example, English. Some clues are more subtle than others and are rare enough to barely surface in text corpora. Finally and most importantly, humans use the physical properties of objects to assign linguistic behavior to words describing these objects. Experiments in different frameworks (one prevalent paradigm being novel word learning) and with different populations have shown that the shape and internal structure of objects, in particular, affect how the corresponding word meaning will be construed (\citealt{soja1991ontological, SAMUELSON19991,PRASADA2002141} a.o.). This, together with data from early cognition, is an argument against Quine's (\citeyear{quine1960word}) strong thesis that language is the instrument for separating the world into objects and substances: some of these distinctions reside in non-linguistic experience, in particular -- visual one (along with other general world knowledge, for example, about how different objects are typically used, see \citealt{MIDDLETON2004371}). This raises the question our paper aims to answer: How will adding visual experience affect individuation, as found in the representations developed by the learner -- in our case, a neural V\&L model?

\section{Individuation Assessment}

We suggest estimating the models' individuation `resolution' by inferring its individuation hierarchy.
The pipeline we construct here is based on publicly available data and is motivated by cognitive and linguistic experiments on individuation discussed above. 

We structure this section as follows. First, we describe the list of nouns 
and semantic features that we use in our experiments throughout this paper. We then propose a simple way to characterize individuation in a model. In short, we will look at embeddings of noun phrases describing various quantities of objects and measure distances between different quantities of the same type of object (that is, described by the same noun). 

While this approach does not, of course, exclude other potential analysis tools, we believe it is a simple and effective way to demonstrate that V\&L models represent individuation differently and show behavior closer to human perception of individuation. We hope that further work on individuation in modern deep neural nets refines the proposed methods or proposes more elaborate ones.

Let us discuss our experiments in detail and then demonstrate how CLIP differs from contextual text embeddings (for example, SBERT, \citealt{reimers2019sentence}) and static word embeddings (say, FastText, \citealt{joulin2016fasttext}) in text-only models. 

\subsection{Data}

We start with a list of nouns alongside their plural forms. We take a publicly available list of singular-plural noun pairs based on an analysis of the Wikipedia corpus\footnote{https://github.com/djstrong/nouns-with-plurals}. This list includes 93 518 words. Since we want to assess individuation across various types of entities, we need to enrich the list with semantic information. We intersect the original list with WordNet\footnote{https://wordnet.princeton.edu/}. This procedure leaves 28 521 nouns from the original list. Now every entry has specific conceptual-semantic attributes from WordNet alongside the plural form of the word that corresponds to a given entry. We use this list for further experiments. 

With more than twenty-eight thousand words, we believe it to be representative and adequate for the broad assessment of individuation in the models we include in the study. For some of the experiments, we filter the obtained list further, leaving several WordNet categories that generally correspond with the taxonomy of individuation hierarchy observed in studies summarized in Section 2. Table \ref{tab:WordNetCat} lists these categories along with the number of words that belong to each category.

\begin{table}[]
\centering
\begin{tabular}{lr}
\hline
\multicolumn{1}{c}{Type} & \multicolumn{1}{c}{Number of Words} \\ \hline
Animal                   & 1887                                \\
Body Part                & 863                                 \\
Fish                     & 220                                 \\
Food                     & 551                                 \\
Fruit                    & 203                                 \\
Living Thing             & 8845                                \\
Nutrient                 & 239                                 \\
Organism                 & 8763                                \\
Person                   & 5861                                \\
Substance                & 1397                                \\
Vascular Plant           & 1027                                \\
Woody Plant              & 470                                
\end{tabular}
\caption{The sizes of the WordNet categories used in the experiments in Subsection \ref{sec:quality}.}
\label{tab:WordNetCat}
\end{table}

\subsection{Embedding Quantities}

In the first experiment, we study quantity distinctions in CLIP. This aspect of the model's behavior is parallel to the `quantity axis' of individuation described in Section 2 in the context of cognition and language. We will compare the model behavior with the results found in humans: we expect that a model that encodes individuation in a manner similar to humans will show starker contrast when comparing smaller quantities (two apples are very different from one apple). In contrast, higher quantities would be less distinguishable (nine apples are very similar to ten apples). We use this logic to construct the following procedure.

\begin{itemize}
\item make a list of phrases following the pattern $n obj_i$, where $n$ is a numeral written in digits, from 2 to 10, and $obj_i$ is a noun from our list in its plural form;
\item calculate embeddings that a chosen model provides with these phrases;
\item for every given noun, calculate pairwise distances for every pair of numeric prefixes;
\item average across nouns and normalize the resulting score.
\end{itemize}

\begin{figure}[]
	\centering
	\begin{subfigure}{.45\textwidth}
		\centering
    	\includegraphics[scale=0.45]{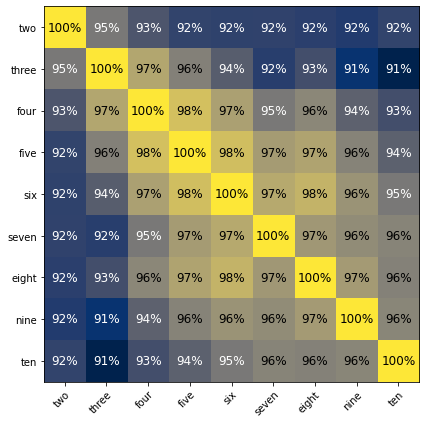}    
    	\caption{CLIP}
    	\label{fig:clip}
	\end{subfigure}\\
	\begin{subfigure}{.45\textwidth}
		\centering
    	\includegraphics[scale=0.45]{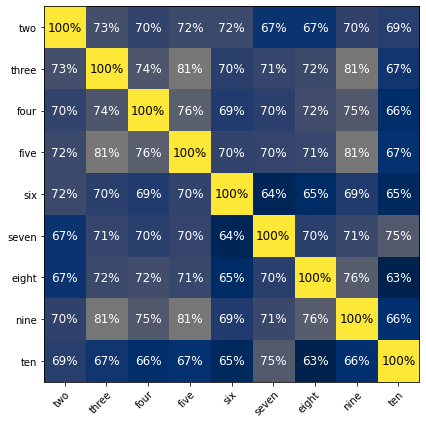}    
    	\caption{FastText}
    	\label{fig:ft}
	\end{subfigure}\\
	\begin{subfigure}{.45\textwidth}
		\centering
        \includegraphics[scale=0.45]{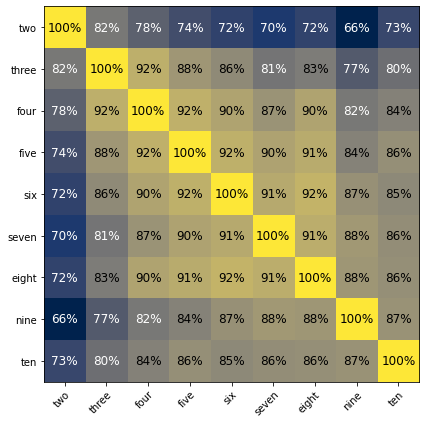}
        \caption{SBERT}
        \label{fig:sbert}
	\end{subfigure}
	\caption{Side by side comparison of contrasting capabilities that models have for various number of objects. The heat map represents average distances for the pairs of embeddings that model provides for various quantities of the same objects. The results are averaged across all objects and normalized.}
	\label{fig:individuation}
\end{figure}

Figure \ref{fig:individuation} illustrates the individuation `resolution' of the models along the quantity axis assessed with the pipeline above. We compare CLIP with FastText \cite{joulin2016fasttext} and SBERT \cite{reimers2019sentence}. Since the scores are normalized, the color scheme of the resulting tables is informative: one should compare the distances between various quantities of entities relative to other quantities. 

\FloatBarrier

Indeed, for FastText, any two different quantities are far apart. SBERT starkly separates two from any other number yet has some difficulties contrasting relatively small numbers, such as three or four, with higher ones, such as nine or ten. Finally, CLIP demonstrates contrasting capabilities that seem closer to the intuition described above. If two numbers are close to each other, the embeddings of the quantities tend to be closer, yet the contrast gets stronger for smaller quantities and weaker for bigger ones.

\subsection{Embedding Qualitative Properties}
\label{sec:quality}

As discussed above, individuation is sensitive to a variety of physical properties of entities, thus giving rise to something we call `the object axis' in Section 2. These properties guide the classification of entities into substances vs. objects. Assessment along this axis enriches the results from the previous section with the other aspect of individuation. 

Figure \ref{fig:individuation} shows contrasts between different quantities averaged across all nouns denoting different types of objects. But the contrast between quantities might decline differently for various classes of nouns, therefore, showing the interaction between the two axes. One can look at the following intuitive example. The individuation scales based on results from cognitive science and linguistics predict that people are higher on those scales (= are more individuateable) than animals or plants. Thus humans might perceive the difference between five and six people as a starker one than the difference between five or six dogs or, say, apples. We suggest seeing whether some of the models in question have similar behavior. 

Since WordNet contains information on the classes of nouns, one could see if the embeddings of the model capture the qualitative properties of the mentioned classes. For this paper, we suggest the following classes:  substance, food, nutrient, body part, vascular plant, woody plant, fruit,  living thing, organism, fish, animal, and person. These are the classes of objects over which we would aggregate the obtained results. Naturally, one could have a less granular picture merging some similar classes, but we suggest using the original WordNet `synset' typology to simplify reproducibility. 

How could one characterize the individuation `resolution' of a given model? As we have mentioned earlier, there is converging cognitive and linguistic data suggesting that individuation and distinguishability of $n$ objects and $n+1$ objects generally declines with higher values of $n$. We have also already shown that all models distinguish two and three objects relatively well. Let us keep these two ideas in mind and introduce a metric that could be a proxy for individuation `resolution.' Let us look at a set of objects: $O = \cup_{k=1}^{k=N} \{ obj_k\}$. Let ${obj_k}^n$ denote $n$ objects, as in {\it I have $n$ apples}, where $obj_k = apple$. Let $M({obj_k}^n)$ denote the embedding that model $M$ has for a noun phrase denoting ${obj_k}^n$. We suggest the following function $I_M$ as the proxy to estimate model $M$ individuation `resolution' for a given object $obj_k$:

\begin{eqnarray}
I_M (obj_k) = \sum_{n=3}^{T} \frac{d(M({obj_k}^n), M({obj_k}^{n+1}))}{T \dot d(M({obj_k}^2), M({obj_k}^{3}))},\nonumber
\end{eqnarray}

where $d(x,y)$ denotes cosine similarity between the corresponding embeddings and $T$ is some finite number. In our experiments, $T=10$. We believe it to be a reasonable assumption\footnote{Most humans would have a hard time differentiating ten and eleven apples on the image without counting them all.}. We want to compare different models in terms of their individuation capabilities, so we need to have some sort of averaging across the objects that our model works with. However, averaging over the whole set $O$ might be too crude. First, we have extensively discussed that humans individuate different classes of objects differently. Second, the embedding spaces of the models might be very different, so there is no reason to believe that two estimates for two different models could be directly compared. However, one could compare values of $I_M$ for a given model on different classes of objects. Say, all objects $obj_j$ belong to a class $C_i$: $C_i = \cup_{j=1}^{j=L} \{ obj_j\}$, then one could introduce an estimator for a given model $M$ on a given class $C$ as follows:

\begin{eqnarray}
\label{eq:ind_class}
&&I_M(C_i) = \frac{\sum_{j=1}^{L} I_M(obj_j)}{L} = \nonumber \\ 
&&\sum_{j=1}^{L} \sum_{n=3}^{T} \frac{d(M({obj_k}^n), M({obj_k}^{n+1}))}{L T \dot d(M({obj_k}^2), M({obj_k}^{3}))}.\nonumber
\end{eqnarray}

The basic intuition behind this metric is that the higher it is, the harder it is for the model to distinguish between $n$ and $n+1$ objects for higher values of $n$. Since all the models in question distinguish two and three objects reasonably well, one could also think of the bigger value for the metric and higher difference between 2 and 3 objects in comparison with higher $n$ and $n+1$ of objects.

Now we can score every class and object with $I_M$ and compare the resulting sets we obtain. We can order different classes of objects $C_i$ in an individuation hierarchy, where classes with lower $I_M(C_i)$ will be placed lower and the classes with higher $I_M(C_i)$ — higher. We can also calculate p-values to characterize to which extent two different classes could be distinguished based on the values of $I_M$. Finally, we can compare the resulting order with the individuation orders suggested for human perception.

Figure \ref{fig:clip_ind} summarizes the resulting orders for various classes $C_i$ and three models: $I_{CLIP}(C_i)$, $I_{SBERT}(C_i)$ and $I_{FastText}(C_i)$. The values in the table are p-values for the relative order of classes with respect to each other. If a value is above 5\%, the difference is not statistically significant.

Now let us discuss the results presented in Figure \ref{fig:clip_ind} and compare the obtained individuation hierarchies with those described in cognitive and linguistic literature.

\section{Discussion}

The first thing that one sees is that FastText has almost no distinguished individuation classes and lumps a variety of classes together. It doesn't give rise to a systematic individuation hierarchy that would be similar in any way to the rankings proposed in cognitive science and linguistics. If anything, the tendency is the opposite: substances turn out to be one of the most `individuated' classes, while animals and organisms end up on the lower end of the scale. 

\begin{figure*}[h!]
	\begin{subfigure}{.9\textwidth}
	\centering
        \includegraphics[scale=0.35]{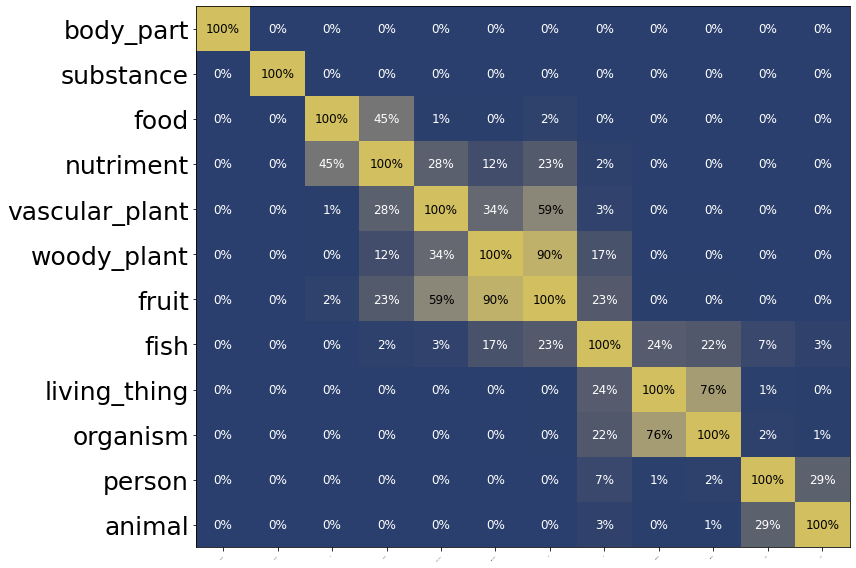}
            	\caption{CLIP}
    	\label{fig:clipind}
	\end{subfigure}\\
	\begin{subfigure}{.9\textwidth}
		\centering
    	\includegraphics[scale=0.35]{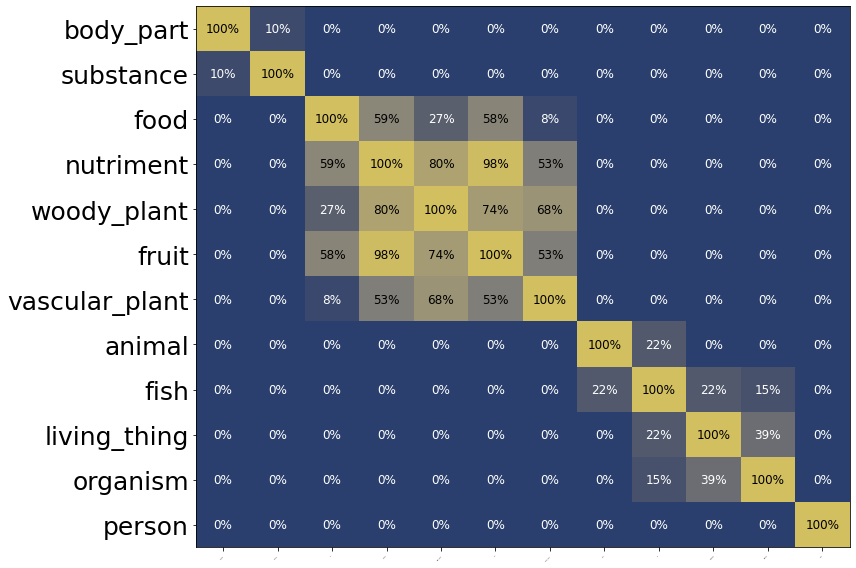}    
    	\caption{SBERT}
    	\label{fig:ftind}
	\end{subfigure}
	\begin{subfigure}{.9\textwidth}
		\centering
    	\includegraphics[scale=0.35]{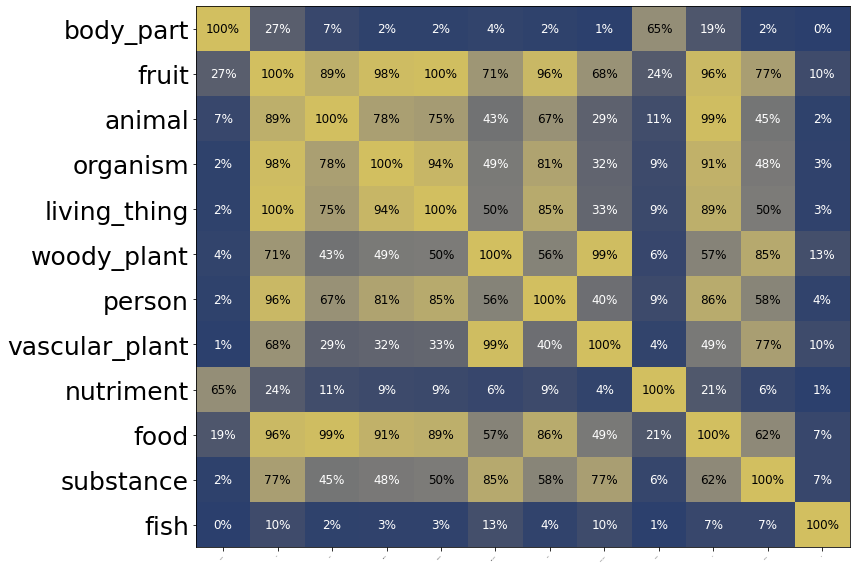}    
    	\caption{FastText}
    	\label{fig:ftind}
	\end{subfigure}
    \caption{P-values for the individuation capabilities of CLIP in comparison with SBERT and FastText based on the proxy metric for individuation. The classes with $p>5\%$ are not significantly distinguishable. The order of rows is in line with the average value of the proposed individuation proxy: the lower individuated classes are on top, the more individuated ones are on the bottom. The order of columns repeats the order of rows making every matrix symmetric.}
    \label{fig:clip_ind}
\end{figure*}

The second observation is that the hierarchies deduced from SBERT and CLIP are similar to the individuation hierarchies proposed in cognitive and linguistic research. As discussed in Section 2, the linguistic behavior of different nouns varies in ways that can be traced back to cognitive distinctions in individuation. Therefore, it comes as no surprise that a text-only model attuned to semantic distinctions that show up in distributional properties of nouns can pick up the relevant contrasts. Yet a careful exploration of Figure \ref{fig:clip_ind} allows us to see where additional -- visual -- modality helps develop clearer individuation-related distinctions compared to the language-only model. 

\FloatBarrier

\begin{table}[t]
\begin{tabular}{lrr}
\hline
\multicolumn{1}{c}{Type} & \multicolumn{1}{c}{\begin{tabular}[c]{@{}c@{}}Number of\\ Separable Cliques\end{tabular}} & \multicolumn{1}{l}{\begin{tabular}[c]{@{}l@{}}Average\\ Clique Size\end{tabular}} \\ \hline
CLIP                     & \textbf{8}                                                                                        & \textbf{2.3}                                                                              \\
SBERT                    & 5                                                                                         & 2.6                                                                                \\
FastText                 & 4                                                                                         & 7.3                                                                             
\end{tabular}
\caption{The parameters of the individuation equivalence graphs by the models. CLIP has the most fine-grained individuation among the compared models.}
\label{tab:cliques}
\end{table}

The individuation scale produced by CLIP is more fine-grained. For example, SBERT lumps fruits in one individuation `cluster' with foods and nutrients and various plants, while CLIP distinguishes foods and nutrients from plants. This can potentially be related to the fact that images depicting food (in particular, fruit) tend to differ from images with plants in general (say, landscapes). One of the ways to quantify the resulting differences in the model's individuation is to represent data shown in Figure \ref{fig:clip_ind} as a graph and calculate the cliques' parameters. Let us connect to vertices representing a class of nouns with an edge if the pval on Figure \ref{fig:clip_ind} is greater than five percent. This would mean that our proxy metric based on model embeddings has difficulty distinguishing the classes. Now we can count maximal cliques. The more cliques we end up with, the more separate classes are distinguished by a given model. We can also calculate the average size of the cliques. The smaller this size is, the more fine-grained the individuation hierarchy induced by a given model is. The results of those calculations are presented in Table \ref{tab:cliques}. Indeed, CLIP provides the most fine-grained individuation hierarchy.

Another interesting aspect of this is the position of animals in the induced individuation hierarchy. While SBERT puts animals somewhere in the middle of the scale, CLIP puts them on top with the same cluster as humans. Such a position goes in line with the cognitive results mentioned earlier in Section 2.

Finally,  going back to Figure \ref{fig:individuation}, one could notice that CLIP individuation has fewer discontinuities when compared to SBERT. Specifically, if $n<k<j$, then CLIP almost always recognizes that $I(n)<I(k)<I(j)$. Out of thirty-six pairwise comparisons, there are three situations when this observation does not hold. With SBERT, this does not hold in eight cases out of thirty-six pairwise comparisons.

Another thing worth noting is the standard deviation of the individuation proxy that differs significantly for all three models; see Table \ref{tab:std}. CLIP shows the lowest standard deviation across categories, while that of FastText is ten times higher. 

\begin{table}[t]
\centering
\begin{tabular}{lr}
\hline
\multicolumn{1}{c}{Type} & \multicolumn{1}{c}{\begin{tabular}[c]{@{}c@{}}Standard\\ Deviation\end{tabular}} \\ \hline
CLIP                     & \textbf{0.016}                                                 \\
SBERT                    & 0.05    \\
FastText                 & 0.12                                                                  
\end{tabular}
\caption{Average standard deviation for the individuation proxy across individuation categories.}
\label{tab:std}
\end{table}

\section{Conclusion}

This paper demonstrates that CLIP benefits from vision-language synergy and thus effectively encodes individuation properties for distinct entities. We suggest a method to calculate whether the model captures individuation for a given class of objects in its embeddings. Using this method, we infer the individuation hierarchy that several models induce on different classes of objects. Thus, we demonstrate that CLIP embeddings capture quantitative differences in a way that is in closer agreement with the human perception of individuation. We hope that this paper stimulates further discussion on multimodality as a source for models that are aligned with human perspective and perception.

\section*{Limitations}

For this research, we used WordNet and CLIP. We believe the results are reproducible with other datasets and V\&L models in languages other than English, but this has not been proved yet. We also use a limited number of models in the comparison. We believe that the observed properties characterize a broader set of multimodal architectures yet restrict our reasoning to CLIP only.

\section*{Ethics Statement}
This paper complies with the \href{https://www.aclweb.org/portal/content/acl-code-ethics}{ACL Ethics Policy}.

\bibliography{anthology,custom}
\bibliographystyle{acl_natbib}





\end{document}